\documentclass[journal]{IEEEtran}

\ifCLASSINFOpdf
\else
\fi

\usepackage{silence}
\usepackage{graphicx}
\usepackage{booktabs}
\usepackage{multirow}
\usepackage{amsmath}
\usepackage{amssymb}
\usepackage{subcaption}
\usepackage{xcolor}
\usepackage[colorlinks, citecolor = blue, urlcolor = magenta]{hyperref} 
\usepackage{algorithm}
\usepackage{listings}
\usepackage[numbers,sort&compress,comma]{natbib}

\begin{document}

\title{Unpaired Image Dehazing via Kolmogorov-Arnold Transformation of Latent Features}

\author{Le-Anh~Tran}

\maketitle

\begin{abstract}
This paper proposes an innovative framework for Unsupervised Image Dehazing via Kolmogorov-Arnold Transformation, termed UID-KAT. Image dehazing is recognized as a challenging and ill-posed vision task that requires complex transformations and interpretations in the feature space. Recent advancements have introduced Kolmogorov-Arnold Networks (KANs), inspired by the Kolmogorov-Arnold representation theorem, as promising alternatives to Multi-Layer Perceptrons (MLPs) since KANs can leverage their polynomial foundation to more efficiently approximate complex functions while requiring fewer layers than MLPs. Motivated by this potential, this paper explores the use of KANs combined with adversarial training and contrastive learning to model the intricate relationship between hazy and clear images. Adversarial training is employed due to its capacity in producing high-fidelity images, and contrastive learning promotes the model's emphasis on significant features while suppressing the influence of irrelevant information. The proposed UID-KAT framework is trained in an unsupervised setting to take advantage of the abundance of real-world data and address the challenge of preparing paired hazy/clean images. Experimental results show that UID-KAT achieves state-of-the-art dehazing performance across multiple datasets and scenarios, outperforming existing unpaired methods while reducing model complexity. The source code for this work is publicly available at \href{https://github.com/tranleanh/uid-kat}{https://github.com/tranleanh/uid-kat}.
\end{abstract}


%
\IEEEpeerreviewmaketitle

\section{Introduction}
\label{sec:intro}

\IEEEPARstart{I}{mage} dehazing is a fundamental challenge in computer vision that aims to improve image clarity degraded by haze, thereby facilitating various important applications such as self-driving cars and surveillance systems \cite{tran2019robust,tran2024toward}. Early prior-based dehazing algorithms, such as DCP \cite{he2010single}, CAP \cite{zhu2015fast}, and BCCR \cite{meng2013efficient}, often rely on empirical observations and specific knowledge of the image formation process. While these methods produce sufficient results in simple scenarios, their reliance on oversimplified statistical assumptions often leads to inadequate performance under real-world complexities. Moreover, these approaches typically require manual parameter tuning, hindering their ease of use and applicability across a wide range of situations.

\begin{figure}
  \centering
  \includegraphics[width=1.0\linewidth]{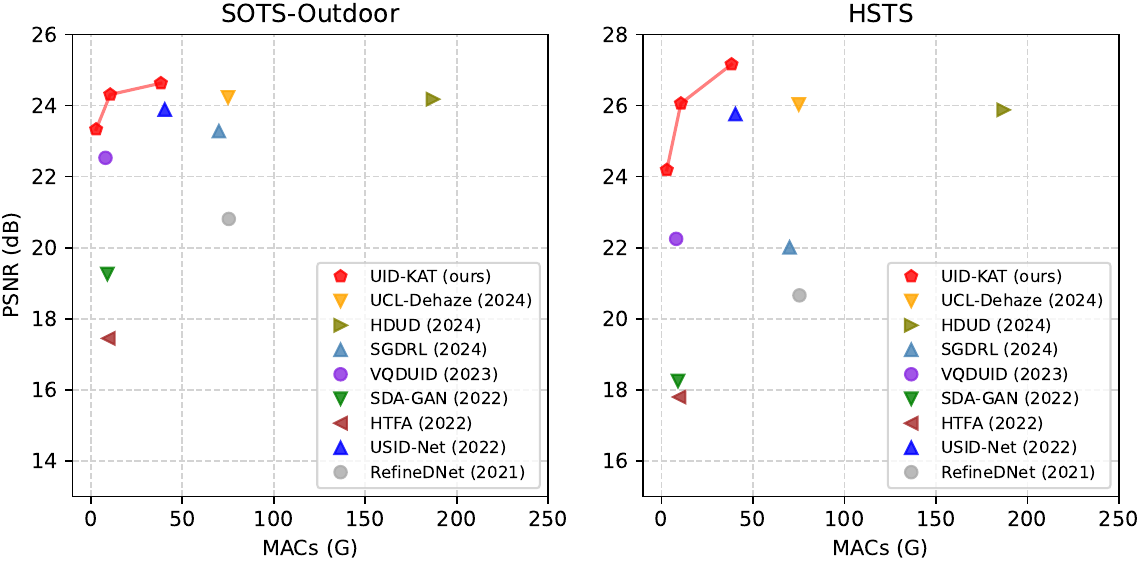}
  \caption{Trade-off between effectiveness (PSNR) and efficiency (MACs) on the SOTS-Outdoor and HSTS benchmarks. The proposed UID-KAT achieves competitive performance with low computational complexity.}
  \label{fig:chart_psnr_macs}
\end{figure}

In the past decade, deep learning-based approaches have emerged as powerful tools for image-to-image translation tasks, with image dehazing being a prominent subfield. The success of these methods can be largely attributed to the flexibility and expressiveness of deep architectures, such as convolutional neural networks (CNNs) and generative adversarial networks (GANs). These models, through full or semi-supervised learning, are able to learn complex mappings between hazy and clear images, achieving superior performance compared to prior-based methods \cite{wu2021contrastive, zhao2021refinednet, tran2022anovel}. However, most conventional learning-based models rely on synthetic paired training data from the same scenes, which are often difficult or even impossible to obtain in real-world conditions. In addition, the gap between synthetic and real-world hazy images inevitably degrades the performance of these models, hindering their generalizability, especially in scenarios with varying haze densities and complex environments. This leads to the introduction of several unsupervised approaches \cite{chen2021psd,li2024hdud,wang2024ucl}. However, unsupervised dehazing remains underexplored, and its performance is still considerably inferior against that of supervised methods. Additionally, most learning-based approaches only exploit clean images as positive samples to guide the network's training, while neglecting the potential value of the information in hazy images, which could serve as negative samples \cite{wang2024ucl}.

In recent years, Vision Transformers (ViTs) \cite{dosovitskiy2020image} have surpassed traditional CNN-based approaches in various vision tasks, challenging the dominance of CNNs in high-level vision applications. Accordingly, numerous ViT-based dehazing models \cite{song2023vision, tran2024distilled} have been introduced, which often rely on multi-layer perceptron (MLP) to mix information across channels. However, MLPs are less effective at optimizing univariate functions and have limited capacity for feature transformation in latent spaces, especially compared to alternatives like splines \cite{yang2024kolmogorov}. This is particularly important in image-to-image translation tasks like dehazing, where latent feature transformation plays a crucial role. Recently, Kolmogorov-Arnold Networks (KANs) \cite{liu2024kan}, based on the Kolmogorov-Arnold representation theorem \cite{schmidt2021kolmogorov}, have been proposed as a potential alternative to MLPs. KANs suggest that any continuous function can be expressed as a finite composition of simpler functions, offering promising capabilities for a range of learning tasks. However, the application of KANs to more advanced tasks, such as image-to-image translation, still remains largely underexplored.

To address the challenges outlined above, this paper proposes an unpaired image dehazing framework inspired by the Domain Transfer Network (DTN) \cite{taigman2022unsupervised}, which can leverage adversarial training to learn a single-direction mapping for robust and effective image dehazing. This approach requires no carefully paired hazy and clean image datasets, allowing the model to be trained on randomly collected image sets. To improve feature representation and transformation in latent space, a novel transformer architecture that replaces MLP layers with KAN layers, termed Dual-GR-KAN Transformer, is introduced. This design fully exploits the power of KAN to enhance the model's expressiveness and performance. Based on current literature, this is the first work that introduces the application of KAN to an unsupervised, ill-posed problem like dehazing and demonstrates favorable results. Furthermore, inspired by Contrastive Unpaired Translation (CUT) \cite{park2020contrastive}, contrastive learning is incorporated to guide the model in focusing on salient features and suppressing irrelevant information. Thanks to CUT, the proposed framework remains effective even when trained with a limited amount of data in both source and target domains. As shown in Fig. \ref{fig:chart_psnr_macs}, the proposed framework, named Unpaired Image Dehazing via Kolmogorov-Arnold Transformation (UID-KAT), demonstrates favorable outcomes in both effectiveness and efficiency, even when trained on only 1,000 images in each domain. In a nutshell, the contributions of the paper are summarized as follows:

\begin{itemize}

  \item This paper proposes UID-KAT, a framework exploring the prowess of KAN for unpaired image dehazing.

  \item This paper introduces the Dual-GR-KAN Transformer module, which fully exploits KAN for feature transformation in the latent space.

  \item Experimental results show that UID-KAT can produce state-of-the-art dehazing performance on various benchmark datasets and natural captured data, surpassing existing methods with lower computational cost.
  
\end{itemize}

The rest of this paper is organized as follows. Section \ref{sec:relatedwork} briefly reviews related research works. In Section \ref{sec:method}, the proposed framework is elaborated. The experiments and analyses on various benchmark datasets are provided in Section \ref{sec:experiments}. Section \ref{sec:conclusions} concludes the paper.

\section{Related Work}
\label{sec:relatedwork}

\subsection{Single Image Dehazing}
\label{subsec:singleimagedehazing}

\textbf{Prior-based Dehazing.} Early image dehazing algorithms heavily rely on hand-crafted priors, which leverage statistical observations about the hazy image formation process to estimate the transmission map and atmospheric light. Dark Channel Prior (DCP) \cite{he2010single}, a well-known representative of traditional method, performs dehazing based on an assumption that the minimum intensity value in a local region of a haze-free image is close to zero in at least one color channel. Color Attenuation Prior (CAP) \cite{zhu2015fast} creates a linear model for modeling the scene depth of the hazy image and learns the parameters of the model via supervised learning. Non-Local Image Dehazing (NLID) \cite{berman2016non} is developed based on an observation that the hazy pixels in the RGB color space are distributed along lines passing through the airlight. Recently, Regional Saturation-Value Translation (RSVT) \cite{tran2024single} is introduced based on statistical analyses of the correlation between hazy points and respective haze-free points in the HSV color space. Although these approaches have yielded impressive dehazing results in controlled environments, they often struggle with scenes containing heavy haze where their simplified assumptions are violated, leading to artifacts and color distortions and hindering their adaptability to real-world scenarios.

\textbf{Learning-based Dehazing.} With notable advancements of deep learning across various vision tasks, many learning-based dehazing methods have been introduced in the literature in the past years. Early efforts such as \cite{cai2016dehazenet,ren2016single,zhang2018densely} have typically focused on predicting transmission map, then the haze-free image is recovered using the atmospheric scattering model. Although these methods laid the foundations for learning-based dehazing, transmission map estimation requires an accurate physical model, which may not be applicable in all scenarios.

To overcome that challenge, some approaches have tried to directly learn the relationship between hazy and haze-free images in an end-to-end manner by using convolutional neural networks (CNNs) \cite{qin2020ffa}, generative adversarial networks (GANs) \cite{dong2020fd,tran2024encoder}, or knowledge distillation \cite{hong2020distilling,tran2024lightweight}. These methods have achieved significant success and effectively provided visually appealing results by training on large-scale datasets of hazy-clean image pairs via supervised learning. Notwithstanding, acquiring such datasets is often challenging and impractical. In addition, supervised methods may struggle to generalize well in unseen environments, especially if the model is overfitted to a specific haze scenario presented in the training data. Recently, these issues have been tackled by several semi-supervised learning studies \cite{li2019semi,an2022semi}, which often combine a small set of paired images with a larger set of unpaired data. While semi-supervised methods present promising solutions and can alleviate domain shift problems to some extent, their performance still depends on the quality and quantity of synthetic data, and they may struggle to generalize effectively to non-ideal conditions, as they rely heavily on the assumptions made by the unpaired data.

Unsupervised learning methods are particularly attractive in scenarios where paired data is unobtainable, such as in the context of image dehazing, making them more versatile in practical scenarios. Some unsupervised dehazing methods, such as \cite{engin2018cycle,li2021you,zhao2021refinednet}, can be easily adapted to various environmental conditions without requiring retraining, since they are not overly dependent on specific haze types. While supervised learning can yield significant performance when environmental constraints are given, unsupervised approaches can offer better generalizability.

\subsection{Vision Transformers}
\label{subsec:vit}

In recent years, the breakthrough of an emerging neural network class, Vision Transformers (ViTs) \cite{dosovitskiy2020image}, has opened up a new research direction for neural architecture design. The abstracted architecture of ViTs has been found to play a significant role in delivering superior performance \cite{yu2022metaformer}. In a general ViT block, an input is first embedded into a sequence of features (or tokens), before being passed through two sub-blocks, roughly termed $\mathrm{TokenMix}(.)$, a token mixing operation which propagates information among tokens, and $\mathrm{ChannelMix}(.)$, which represents an MLP module to fuse features among channels. Various studies \cite{guo2022image,song2023vision,tran2024distilled} have also investigated ViT-based architectures for image dehazing and achieved promising results. However, these works have primarily focused on using ViTs for supervised dehazing, leaving their potential in unsupervised dehazing largely unexplored.

\begin{figure*}
    \centering
    \includegraphics[width=1.00\textwidth]{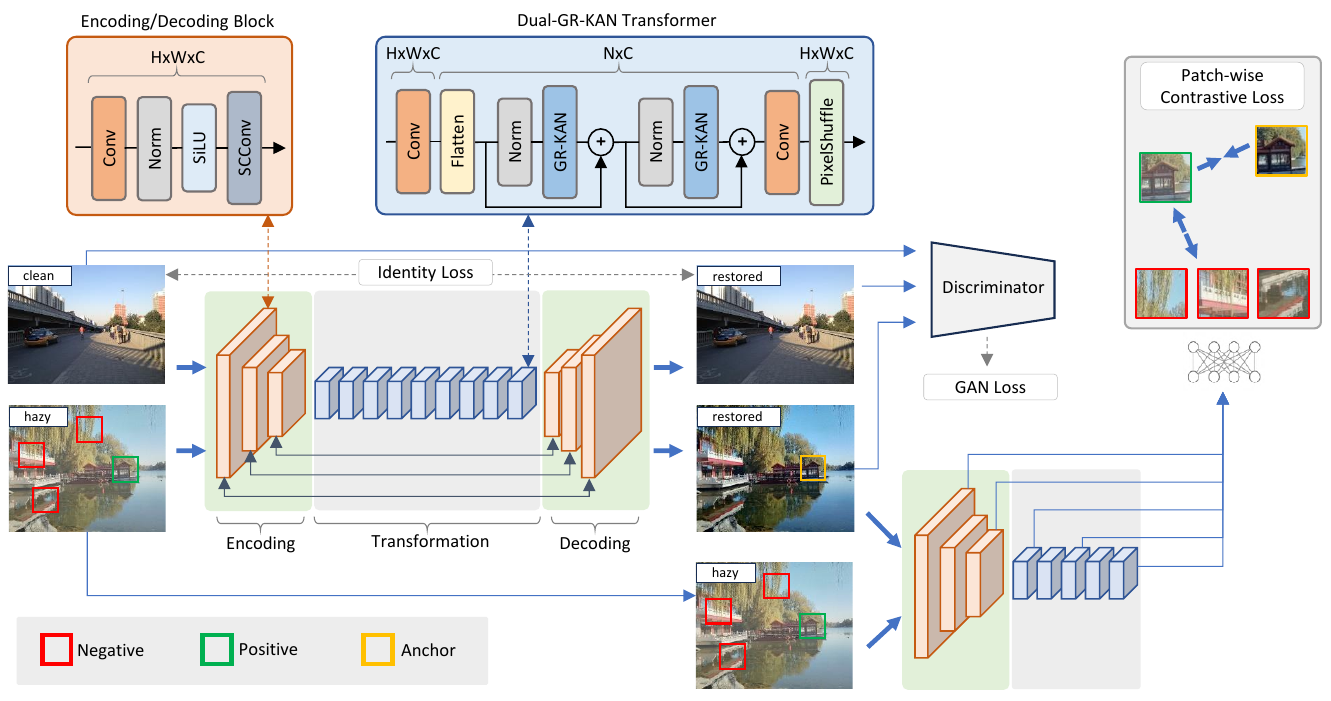}
    \caption{The proposed UID-KAT framework.}
    \label{fig:framework} 
\end{figure*}

\subsection{Contrastive Learning}
\label{subsec:contrastivelearning}

Contrastive learning has emerged as a powerful self-supervised learning technique, leveraging the idea of contrasting positive samples against negative ones. Contrastive learning can encourage the model to learn generalizable and meaningful representations of data in scenarios where acquiring paired data is challenging, which is particularly useful in the context of image dehazing. Inspired by this, several works such as \cite{wu2021contrastive,wang2024ucl} have examined contrastive learning for image dehazing and achieved promising outcomes. Particularly, another work \cite{wang2023uscformer} even incorporates contrastive learning and ViTs in an supervised manner to obtain impressive results. However, applying contrastive learning to image dehazing still presents challenges as it can be difficult to achieve effective positive/negative pair selection for successful training, especially in unsupervised settings. Albeit these difficulties, the potential of contrastive learning to advance unsupervised image dehazing methods remains significant, particularly in its ability to learn robust features that can aid in haze removal even when paired data is unavailable.

\subsection{Kolmogorov-Arnold Networks (KANs)}
\label{subsec:kan}

The Kolmogorov-Arnold representation theorem \cite{schmidt2021kolmogorov}, established by Vladimir Arnold and Andrey Kolmogorov, states that any multivariate continuous function $f$ on a bounded domain can be expressed as a finite composition of continuous functions of a single variable and the binary operation of addition \cite{liu2024kan}. Specifically, for a smooth $f: [0,1]^n \to \mathbb{R}$:
\begin{equation}
f(\mathbf{x}) = f(x_1, \ldots, x_n) = \sum_{q=1}^{2n+1} \Phi_q\left(\sum_{p=1}^{n} \phi_{q,p}(x_p)\right),
\end{equation}
where $\phi_{q,p}: [0,1] \to \mathbb{R}$ and $\Phi_q: \mathbb{R} \to \mathbb{R}$. The theorem can be written in matrix form as follows:
\begin{equation}
    f(\mathbf{x}) = \Phi_{\text{out}} \circ \Phi_{\text{in}} \circ \mathbf{x} \label{eq:kan_vector},
\end{equation}
where $\Phi_{\text{in}}$ and $\Phi_{\text{out}}$ are defined as:
\begin{equation}
    \Phi_{\text{in}} = \begin{bmatrix}
\phi_{1,1}(\cdot) & \cdots & \phi_{1,n}(\cdot) \\
\vdots & \ddots & \vdots \\
\phi_{2n+1,1}(\cdot) & \cdots & \phi_{2n+1,n}(\cdot)
\end{bmatrix},
\end{equation}
\begin{equation}
    \Phi_{\text{out}} = \begin{bmatrix}
\Phi_1(\cdot) & \cdots & \Phi_{2n+1}(\cdot)
\end{bmatrix}.
\end{equation}
This breakdown shows how $f$ can be constructed from simpler functions, highlighting a key characteristic of multivariate continuous functions. Inspired by this, a generalized Kolmogorov-Arnold layer with $d_\text{in}$-dimensional inputs and $d_\text{out}$-dimensional outputs is proposed in \cite{yang2024kolmogorov} as an activation function form to learn univariate functions on edge, which is illustrated as:
\begin{equation}
    f(\mathbf{x}) = \Phi \circ \mathbf{x} = \begin{bmatrix}\sum_{i=1}^{d_{in}} \phi_{i,1}(x_i) & \dots& \sum_{i=1}^{d_{in}} \phi_{i,d_{out}}(x_i)\end{bmatrix}, \label{eq:kan_layer}
\end{equation}
where
\begin{equation}
\Phi = \begin{bmatrix}
\phi_{1,1}(\cdot) & \cdots & \phi_{1,d_{\text{in}}}(\cdot) \\
\vdots & \ddots & \vdots \\
\phi_{d_{\text{out}},1}(\cdot) & \cdots & \phi_{d_{\text{out}},d_{\text{in}}}(\cdot)
\end{bmatrix}.
\end{equation}
Note that Eq. (\ref{eq:kan_layer}) can be considered a generalized form of Eq. (\ref{eq:kan_vector}), such that $\Phi = \Phi_{\text{in}}\circ \Phi_{\text{out}}$. In \cite{liu2024kan}, $\Phi$ is parameterized by using a linear combination of SiLU activation \cite{elfwing2018sigmoid} and a B-spline function:
\begin{equation}
        \phi(x) = w_b \mathrm{silu}(x) + w_s \mathrm{spline}(x),
\end{equation}
where
\begin{equation}
\mathrm{silu}(x) = \frac{x}{1+e^{-x}}, 
\end{equation}
\begin{equation}
\mathrm{spline}(x)=\sum_i c_i B_i(x).
\end{equation}
The adaptive activation function nature enables KANs to effectively approximate non-linear mappings with fewer parameters and improved interpretability. Following \cite{liu2024kan}, various subsequent studies have also investigated the promise of KANs in different aspects. Notable research works on KANs can be witnessed in KAT \cite{yang2024kolmogorov}, which integrates KAN layer into the ViT architecture, or KAN-CUT \cite{mahara2024dawn}, which represents the initial attempt of applying KANs to image generation. Despite their promising capabilities, KANs remain in their early stages, with limited research investigating their applications across diverse domains. One contributing factor may be the scalability constraints of KANs \cite{yang2024kolmogorov}. Nevertheless, the potential of KANs to model complex relationships in image data has not yet been fully explored, positioning a promising approach for numerous vision-related challenges, particularly in image restoration tasks such as image dehazing.

\section{Methodology}
\label{sec:method}

\subsection{Overview}
\label{subsec:overview}

The proposed framework leverages adversarial learning and feature-level consistency to enhance dehazing performance in an unpaired setting. An overview of the framework is depicted in Fig. \ref{fig:framework}. The generator translates hazy images into haze-free outputs through three primary stages: encoding, latent feature transformation, and decoding. Initially, the encoder extracts salient features from the input image, generating a feature representation. The latent feature transformation stage is then carried out by nine sequential units of the Dual-GR-KAN Transformer (detailed in Section \ref{subsec:architectures}), which refine and transform features from the ``hazy" domain to the ``clean" domain. Finally, the decoder reconstructs the haze-free image utilizing the refined features, with skip connections established between the encoder and decoder for preserving low-level information. To ensure semantic fidelity and structural consistency, patch-level contrastive learning is employed which helps to maximize mutual information between corresponding patches in a learned feature space. The networks are trained using a combination of multiple loss functions, collectively enhancing visual quality, structural detail, and semantic coherence in the generated images. By integrating the adversarial learning capabilities of GANs with the feature transformation strengths of KANs, the proposed framework delivers robust dehazing performance, producing visually compelling and structurally consistent haze-free images.

\subsection{Architectures}
\label{subsec:architectures}

\textbf{Dual-GR-KAN Transformer.} The abstracted architecture of ViTs plays a pivotal role in their success across various vision tasks \cite{yu2022metaformer}. Building on this foundational insight and leveraging the robust feature representation capabilities of KANs, a ViT-based architecture with integrated KAN layers is designed to enhance latent feature transformation in image translation tasks, particularly for unpaired dehazing. Although previously similar efforts have focused on specific domains \cite{VisionKAN2024,chen2024sckansformer}, these methods often fall short in achieving scalability for more complex tasks when combining KANs and ViTs. Recently, Group-Rational KAN (GR-KAN) \cite{yang2024kolmogorov}, a novel variant of KANs, has gained significant attention. This variant replaces the B-spline basis with a rational function and incorporates parameter sharing across groups of edges, resulting in enhanced performance and scalability. Specifically, the function $\phi(x)$ for each edge is parameterized as a rational function of polynomials $P(x)$ and $Q(x)$ of order $m$ and $n$, respectively:
\begin{equation}
\phi(x) = wF(x) = w\frac{P(x)}{Q(x)} = w\frac{a_0 + a_1x+\dots + a_m x^m}{b_0 + b_1x+\dots + b_n x^n},
\end{equation}
where $a_n$ and $b_m$ denote the rational function coefficients, and $w$ indicates the scaling factor. The degree of this function is defined as $m/n$, and $a_n$, $b_m$, $w$ are learned through end-to-end backpropagation. To address potential instabilities caused by poles (situations where the denominator $Q(x)\to 0$ leads to $\phi(x)\to \pm\infty$), a Safe Padé Activation Unit is employed, which is specifically designed to mitigate instability by preventing the function from becoming unbounded:
\begin{equation}
F(x) = \frac{a_0 + a_1x+\dots + a_m x^m}{1 + |b_1x+\dots + b_n x^n|}. \label{eq:rational}
\end{equation}
Building upon the demonstrated effectiveness of GR-KAN \cite{yang2024kolmogorov}, it is incorporated into a ViT-like architecture to enhance latent feature transformation. Specifically, GR-KAN is integrated within both the $\mathrm{TokenMix}(.)$ and $\mathrm{ChannelMix}(.)$ operations to facilitate effective feature refinement. Given an input feature map of shape $\mathbb{R}^{H \times W \times C}$ and a predefined patch size $P$, a patch embedding process is first applied to partition the input into smaller patches. These patches are then flattened into a feature sequence of shape $\mathbb{R}^{N \times C}$, where $ N = (H/P) \times (W/P) $. This stage is implemented using a convolutional layer with a kernel size of $P$ and a stride of $P$, followed by a flattening operation. The flattened feature sequence then undergoes sequential transformation through two identical processing blocks, each consisting of a normalization layer followed by a GR-KAN layer. To ensure stable gradient flow and mitigate potential information loss, a residual connection is employed within each block. A patch unembedding stage afterward is carried out to reshape the transformed feature sequence back into its original dimensions, $\mathbb{R}^{H \times W \times C}$. This is achieved by using a convolutional layer followed by a subpixel operation, effectively reconstructing spatial structures while preserving feature integrity. The resulting module, termed the Dual-GR-KAN Transformer, is a highly versatile component suitable for a wide range of image-to-image translation tasks and multi-scale feature fusion. Given that feature maps in such tasks are typically represented in the form $\mathbb{R}^{H \times W \times C}$, maintaining this format is crucial for preserving fine-grained information and ensuring spatial coherence. Additionally, retaining the $\mathbb{R}^{H \times W \times C}$ representation facilitates the patch-wise contrastive learning employed in the framework. For the default configuration, the patch size is set to $4 \times 4$. The effects of different patch sizes are also analyzed in the ablation study. The straightforward yet effective design of the Dual-GR-KAN Transformer is illustrated in Fig. \ref{fig:framework}.

\textbf{Generator.} As illustrated in Fig. \ref{fig:framework}, the generator is designed by introducing several key modifications to the standard ResNet-based architecture, which serves as the baseline and is widely employed in unpaired image-to-image translation tasks \cite{johnson2016perceptual,zhu2017unpaired,wang2024ucl}. The baseline architecture comprises a $4 \times$ downscaling encoder, followed by nine residual blocks, and a $4 \times$ upscaling decoder. To capture long-range spatial dependencies and inter-channel interactions at each spatial location as well as expanding the receptive field, the self-calibration convolution (SCConv) module \cite{liu2020improving} is incorporated into every stage of both the encoder and decoder. Furthermore, ReLU activations are replaced with SiLU activations, as SiLU has been shown to provide smoother gradients and improved performance, particularly in complex image restoration tasks \cite{elfwing2018sigmoid}. While various activation functions were assessed in the ablation study, SiLU consistently outperformed the alternatives. A major architectural enhancement involves replacing the nine conventional residual blocks in the latent feature transformation stage with nine Dual-GR-KAN Transformer modules, which significantly enhance feature refinement. Additionally, to preserve low-level information and mitigate gradient vanishing issues, skip connections are incorporated between the encoder and decoder. As summarized in Table \ref{table:modelversions}, three model variants \{tiny, small, base\} with respective computational costs of \{2.88, 10.59, 38.28\} GMACs are presented to accommodate different computational constraints. These variants are substantially more compact than existing models in the literature (refer to Table~\ref{table:resultsonsotshsts} for detailed comparisons). While larger network configurations could potentially yield improved quantitative results, the primary objective of this research is to design computationally efficient models. Consequently, the base model represents the largest configuration considered in this study.

\textbf{Discriminator.} The discriminator architecture is inspired by the widely adopted PatchGAN model \cite{isola2017image}. Unlike traditional discriminators that make a single global decision regarding the authenticity of an entire image, PatchGAN divides the image into smaller patches and evaluates each patch independently. This patch-level assessment enables the discriminator to focus on local image details, leading to more robust performance in distinguishing real from generated content. PatchGAN operates in a fully convolutional manner, allowing it to process images of arbitrary sizes without requiring a fixed input dimension. This design not only enhances flexibility but also reduces the overall number of parameters compared to full-image discriminators, resulting in faster processing speeds and improved computational efficiency. Due to its lightweight architecture and proven effectiveness, PatchGAN has become a foundational component in numerous GAN-based frameworks \cite{isola2017image,zhu2017unpaired}, playing a critical role in advancing tasks such as image synthesis and translation.

\begin{table}[t]
\centering
\caption{Variants of UID-KAT: tiny (T), small (S), base (B). All variants use a patch size of $4 \times 4$ and an embedding size of $256$.}
\resizebox{0.48\textwidth}{!}{
\begin{tabular}{ccccc}
\toprule
Variant & \#Channels(1st) & \#Blocks & \#Param(M) & MACs(G) \\
\midrule
UID-KAT-T & 16 & 9 & 1.94 & 2.88 \\
UID-KAT-S & 32 & 9 & 7.70 & 10.59 \\
UID-KAT-B & 64 & 5 & 18.08 & 38.28 \\
\bottomrule
\end{tabular}}
\label{table:modelversions}
\end{table}

\begin{table*}[ht]
\centering
\caption{Performance comparisons on the SOTS-Outdoor and HSTS datasets.}
\begin{tabular}{llcccccrr}
\toprule
\multirow{2}{*}{\text{Method}} & \multirow{2}{*}{\text{Publication}} & \multirow{2}{*}{\text{Type}} & \multicolumn{2}{c}{\text{SOTS-Outdoor}} & \multicolumn{2}{c}{\text{HSTS}} & \multicolumn{2}{c}{\text{Overhead}} \\
\cmidrule(r){4-5} \cmidrule(r){6-7} \cmidrule(r){8-9}
& & & \text{PSNR$\uparrow$} & \text{SSIM$\uparrow$} & \text{PSNR$\uparrow$} & \text{SSIM$\uparrow$} & \text{\#Param(M)} & \text{MACs(G)} \\
\midrule
DCP \cite{he2010single} & TPAMI'11 & Prior & 18.38 & 0.819 & 17.01 & 0.803 & - & - \\
BCCR \cite{meng2013efficient} & ICCV'13 & Prior & 15.71 & 0.769 & 15.21 & 0.747 & - & - \\
JEEF \cite{kaplan2023real} & JVCIR'23 & Prior & 17.51 & 0.857 & 17.61 & 0.861 & - & - \\
RSVT \cite{tran2024single} & PCS'24 & Prior & 22.31 & 0.903 & 21.53 & 0.887 & - & - \\
\midrule
ZID \cite{li2020zero} & TIP'20 & Zero-shot & 20.27 & 0.878 & 22.65 & 0.901 & 40.41 & 1.18 \\
YOLY \cite{li2021you} & IJCV'21 & Unsupervised & 20.39 & 0.889 & 21.02 & 0.905 & 32.00 & * \\
PSD \cite{chen2021psd}	& CVPR'21 & Unsupervised & 20.49 & 0.844 & 19.37 & 0.824 & 33.11 & 182.50 \\
RefineDNet \cite{zhao2021refinednet} & TIP'21 & Weakly-supervised & 20.81 & 0.877 & 20.66 & 0.890 & 65.80 & 75.41 \\
USID-Net \cite{li2022usid} & TMM'22 & Unsupervised & 23.89 & \textcolor{blue}{0.919} & 25.76 & \textcolor{blue}{0.923} & 3.77 & 40.42 \\
HTFA \cite{li2022haze} & KBS'22 & Unsupervised & 17.45 & 0.704 & 17.98 & 0.695 & 1.57 & 9.63 \\
SDA-GAN	\cite{dong2022semi} & TransCyb'22 & Semi-supervised & 19.26 & 0.813 & 18.08 & 0.776 & 19.96 & 9.02 \\
VQDUID \cite{yang2023visual} & NN'23 & Unsupervised & 22.53 & 0.875 & 22.25 & 0.847 & 0.23 & 7.98 \\
SGDRL \cite{jia2024self} & NN'24 & Unsupervised & 23.28 & \textbf{0.920} & 22.01 & 0.888 & 1.91 & 68.75 \\
HDUD \cite{li2024hdud} & NCAA'24 &Unsupervised & 24.18 & 0.869 & 25.88 & 0.922 & 15.06 & 187.50 \\
UCL-Dehaze \cite{wang2024ucl} & TIP'24 & Unsupervised & 24.23 & 0.914 & 26.03 & 0.919 & 19.45 & 78.85 \\
\midrule
UID-KAT-T & Proposed & Unsupervised & 23.33 & 0.909 & 24.19 & 0.909 & 1.94 & 2.88 \\
UID-KAT-S & Proposed & Unsupervised & \textcolor{blue}{24.31} & 0.917 & \textcolor{blue}{26.06} & 0.922 & 7.70 & 10.59 \\
UID-KAT-B & Proposed & Unsupervised & \textbf{24.57} & 0.917 & \textbf{27.16} & \textbf{0.930} & 18.08 & 38.28 \\
\bottomrule
\end{tabular}
\label{table:resultsonsotshsts}
\end{table*}

\subsection{Loss Functions}
\label{subsec:contrastive}

\textbf{Least Squares Loss.} LSGANs \cite{mao2017least} were proposed to overcome the vanishing gradient problem often encountered in traditional GANs. By mitigating this issue, LSGANs facilitate the generation of higher-quality images and improve the stability of the training process. The least squares loss function to train the models is defined as:
\begin{equation}\label{eq:advg}
\begin{aligned}
\mathcal{L}_{Adv}(G)=E_{G(x) \sim P_{fake}}[(D(G(x))-1)^2],
\end{aligned}
\end{equation}
\begin{equation}\label{eq:advd}
\begin{aligned}
\mathcal{L}_{Adv}(D)=E_{y \sim P_{real}}[(D(y)-1)^2]\\
+E_{G(x) \sim P_{fake}}[(D(G(x)))^2],
\end{aligned}
\end{equation}
where $D(.)$ indicates the discriminator's output, while $G(.)$ and $y$ denote the restored and clean images, respectively.

\textbf{Identity Loss.} The identity loss function is adopted to prevent the generator from making unnecessary modifications or over-processing the input image when it is clean, which could result in the loss of important details or the introduction of unwanted artifacts \cite{wang2024ucl}. By penalizing such deviations, the identity loss encourages the model to preserve the intrinsic characteristics and fine details of the input image. The identity loss is defined as:
\begin{equation}\label{identityloss}
\begin{array}{l}
\mathcal{L}_{Ide}=E_{y \sim P_{data(Y)}}\left[\left\|\left(G({y})-y\right)\right\|_{1}\right].
\end{array}
\end{equation}

\textbf{Patch-wise Contrastive Loss.} Inspired by \cite{park2020contrastive,wang2024ucl}, contrastive learning is applied to bring the embeddings of corresponding patches closer within a learned feature space, thereby maximizing mutual information. Recognizing that both the encoder and feature transformation blocks are trained to generate hidden image representations, the feature stack they produce is readily accessible for this purpose. Each representation within this stack represents a spatial location patch of the input image, with deeper layers corresponding to larger patches. Specifically, $N+1$ patches are randomly selected from the input image $x$, and one corresponding patch is chosen from the restored image $y$, as shown in Fig. \ref{fig:framework}. Thanks to the design of the Dual-GR-KAN Transformer which yields feature maps with shape $\mathbb{R}^{H \times W \times C}$, this feature selection process can easily be carried out. In this context, the pair of corresponding patches from $x$ and $y$ constitutes the positive samples, while the remaining $N$ patches from $x$ serve as negative samples. Given that the anchor patch (from $y$), the positive patch, and the $N$ negative patches are denoted as $v$, $v^+$, and $v^-$, respectively, a noisy contrastive estimation module is then employed to maximize the mutual information between the positive samples, formulating an ($N+1$)-way classification problem \cite{park2020contrastive}, with the probability of selecting $v^+$ over $v^-$ computed as follows:
\begin{equation}\label{patchNCE}
\begin{array}{l}
\ell\left({v}, {v}^{+}, {v}^{-}\right)= \\
-\log \left(\frac{\exp \left({\mathrm{sim}}\left(v, {v}^{+}\right) / \tau\right)}{\exp \left({\mathrm{sim}}\left(v, {v}^{+}\right) / \tau\right)+\sum_{n=1}^{N} \exp \left({\mathrm{sim}}\left(v, {v}_{n}^{-}\right) / \tau\right)}\right),
\end{array}
\end{equation}
where $\mathrm{sim}(u,v)$ denotes the cosine similarity between $u$ and $v$, while $\tau$ ($\tau=0.07$) is a temperature parameter to adjust the distance between the anchor and other samples \cite{park2020contrastive}.

\begin{figure*}
    \centering
    \includegraphics[width=0.98\textwidth]{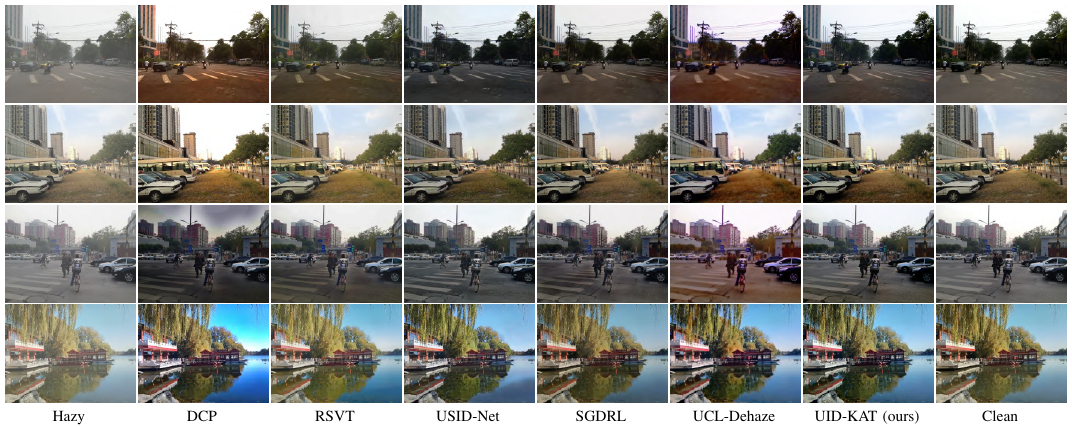}
    \caption{Typical visual results on SOTS-Outdoor.}
    \label{fig:sotsresults} 
\end{figure*}

\begin{figure*}
    \centering
    \includegraphics[width=0.98\textwidth]{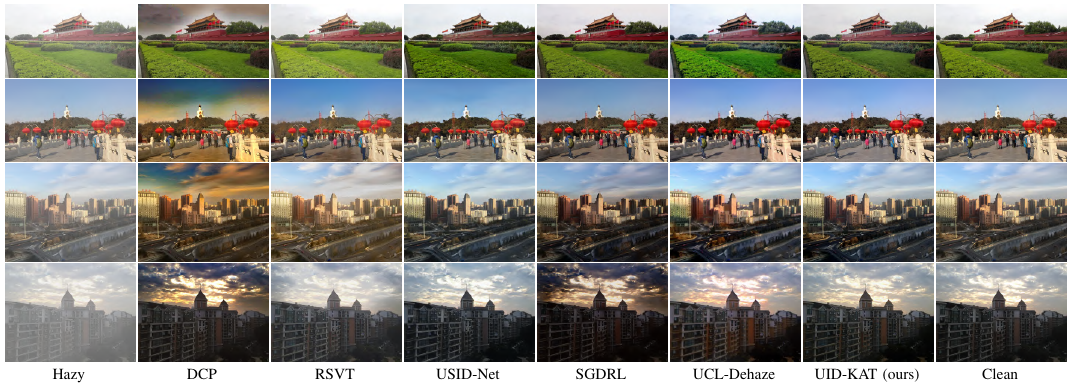}
    \caption{Typical visual results on HSTS.}
    \label{fig:hstsresults} 
\end{figure*}

Accordingly, multiple layers of interest are selected and the corresponding feature maps are passed through a two-layer MLP network, denoted as $H$, to extract features from the input image $x$, yielding a stack of features, represented as $\left\{z_{l}\right\}_{L}=\left\{H^{l}\left(G_{enc}^{l}(x)\right)\right\}_{L}$, where $L$ refers to the number of selected layers from the encoder $G_{enc}$, and $l$ indexes the specific $l$-th chosen layer. These feature stacks correspond to different patches of the image, with each selected layer containing spatial locations indexed by $s \in\left\{1, \ldots, S_{l}\right\}$, where $S_{l}$ denotes the number of spatial locations in layer $l$. For each selection, an anchor patch is chosen, and its feature is denoted as $\hat{z}_{l}^{s} \in \mathbb{R}^{C_{l}}$, where $C_{l}$ represents the number of channels in the respective layer. The positive sample is denoted as ${z}_{l}^{s} \in \mathbb{R}^{C_{l}}$, while the negative samples are represented as ${z}_{l}^{S \backslash s} \in \mathbb{R}^{\left(S_{l}-1\right) \times C_{l}}$. Consequently, the patch-wise contrastive loss is formulated as:
\begin{equation}\label{patchNCE_2}
\begin{array}{l}
{L}_{PC}(G, H, X)=\mathbb{E}_{{x} \sim X} \sum_{l=1}^{L} \sum_{s=1}^{S_{l}} \ell\left(\hat{z}_{l}^{s}, {z}_{l}^{s}, {z}_{l}^{S \backslash s}\right).
\end{array}
\end{equation}

\textbf{Integrated Loss.} The final integrated loss function is formulated by combining all the aforementioned loss functions using balancing weights:
\begin{equation}\label{integrated}
\mathcal{L}_{final} = \lambda_1 \mathcal{L}_{Adv}(G) + \lambda_2 \mathcal{L}_{Ide} + \lambda_3 \mathcal{L}_{PC},
\end{equation}
where balancing weights are set as follows: $\lambda_1 = 1$, $\lambda_2 = 1$, and $\lambda_3 = 5$, as inherited from \cite{wang2024ucl}.

\section{Experiments}
\label{sec:experiments}

\subsection{Experimental Details}
\label{subsec:settings}

\textbf{Datasets.} The proposed UID-KAT framework is trained in an unsupervised learning paradigm, which leverages the abundance of real-world data and simplifies the data preparation process. In accordance with established methodologies \cite{zhao2021refinednet,wang2024ucl}, various publicly available datasets from the RESIDE database \cite{li2018benchmarking} have been utilized for training the models. Specifically, 1,000 hazy images were randomly selected from three distinct RESIDE's subsets \{OTS, RTTS, and URHI\} for the source domain. For the target domain, a set of 1,000 clean images was sampled from two subsets \{OTS and ITS\}. The SOTS-Outdoor and HSTS datasets have been adopted for evaluation and comparison, with 500 and 10 corresponding hazy/clean outdoor image pairs, respectively. Additionally, the generalization capability of the proposed networks in real-world scenarios has been assessed using a separate collection of natural hazy image data.

\textbf{Comparison Methods.} A comparative analysis has been conducted to compare UID-KAT against several state-of-the-art approaches, which are categorized into two main classes: prior-based algorithms and deep learning models. To ensure a fair comparison, supervised methods are excluded from the comparison, as UID-KAT operates in an unpaired setting. The selected DL methods include unsupervised, weakly-supervised, semi-supervised, and zero-shot approaches, ensuring a fair evaluation within the context of unsupervised learning. The prior-based techniques considered in this study include DCP~\cite{he2010single}, BCCR~\cite{meng2013efficient}, JEEF~\cite{kaplan2023real}, and RSVT~\cite{tran2024single}. The DL-based methods include recent state-of-the-art models, such as ZID~\cite{li2020zero}, YOLY~\cite{li2021you}, PSD~\cite{chen2021psd}, RefineDNet~\cite{zhao2021refinednet}, USID-Net~\cite{li2022usid}, HTFA~\cite{li2022haze}, SDA-GAN~\cite{dong2022semi}, VQDUID~\cite{yang2023visual}, SGDRL~\cite{jia2024self}, HDUD~\cite{li2024hdud}, and UCL-Dehaze~\cite{wang2024ucl}. These methods represent the current advancements in the field, providing a robust basis for assessing the performance of the proposed dehazing scheme.

\textbf{Experimental Settings.} The experiments were conducted on a Linux operating system with an Intel(R) Xeon(R) Gold 6134 @ 3.20GHz CPU and GeForce GTX TITAN X GPUs. The proposed framework was implemented using PyTorch, with the Adam optimizer~\cite{kingma2014adam} utilized for training. The training process was conducted with a batch size of 1 over a total of 100 epochs. The initial learning rate was set to \(2 \times 10^{-4}\) for the first 50 epochs and was linearly reduced to zero over the remaining 50 epochs. The networks were trained with an input resolution of \(256 \times 256\). The quantitative evaluation on synthetic datasets was carried out using Peak Signal-to-Noise Ratio (PSNR) and Structural Similarity Index Measure (SSIM) to assess reconstruction quality, whereas the computational complexity was analyzed based on the number of parameters (\#Params) and multiply-accumulate (MAC) operations.

\begin{figure*}
    \centering
    \includegraphics[width=0.98\textwidth]{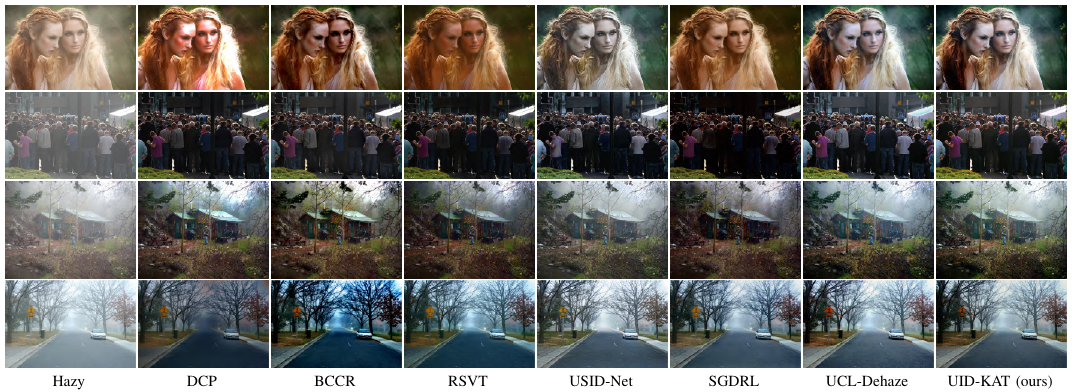}
    \caption{Typical visual results on natural data.}
    \label{fig:naturalresults} 
\end{figure*}

\subsection{Performance Evaluations}
\label{subsec:evaluations}

\textbf{Quantitative Evaluations.} The quantitative comparisons presented in Table \ref{table:resultsonsotshsts} highlight the effectiveness of the proposed UID-KAT models on the SOTS-Outdoor and HSTS datasets. As compared to prior-based methods like DCP and BCCR, the proposed models demonstrate substantial improvements. Generally, UID-KAT-B achieves the best performance among all methods, with a PSNR of 24.57 dB and an SSIM of 0.917 for the SOTS-Outdoor dataset, and a PSNR of 27.16 dB and an SSIM of 0.930 for the HSTS dataset. This basically surpasses other recent methods such as SGDRL, HDUD, and UCL-Dehaze while maintaining competitive overhead, particularly in MACs. Notably, UID-KAT-S also achieves favorable results, with a PSNR of 24.31 dB and an SSIM of 0.917 on SOTS-Outdoor, and a PSNR of 26.06 dB and an SSIM of 0.922 on HSTS. Additionally, the tiny model, UID-KAT-T, achieves notable performance with reduced overhead (1.94M parameters and 2.88G MACs), striking a balance between effectiveness and efficiency. These results confirm the superiority of the UID-KAT models in dehazing tasks for both quantitative metrics and computational complexity.

\textbf{Qualitative Evaluations.} The qualitative comparisons of dehazing results on the SOTS-Outdoor and HSTS datasets are illustrated in Fig. \ref{fig:sotsresults} and Fig. \ref{fig:hstsresults}, respectively. In Fig. \ref{fig:sotsresults}, UID-KAT consistently restores vibrant colors and fine details that are either over-smoothed or inaccurately reconstructed by other methods. For instance, in the first and third rows, which show complex urban scenes, UID-KAT effectively preserves the structural details of the roads, buildings, and vehicles while ensuring a perceptually accurate color balance. In contrast, other learning-based models such as SGDRL and UCL-Dehaze fail to recover sufficient details, while traditional methods such as DCP and RSVT often result in over-saturated outcomes. Similarly, in Fig. \ref{fig:hstsresults}, UID-KAT outperforms competing methods by delivering visually realistic results, successfully eliminating haze while retaining rich textures and producing results closer to the ground truth data. 

Additionally, the qualitative comparisons on some natural image data are presented in Fig. \ref{fig:naturalresults}, showcasing that UID-KAT successfully reconstructs fine details, remarkable clarity, and color accuracy of the subjects. In contrast, prior-based algorithms such as DCP and BCCR produce outputs with color distortion, while other learning-based approaches like SGDRL fail to restore the image's natural vibrancy effectively. As a result, UID-KAT demonstrates robust performance by achieving natural and balanced reconstructions, free from undesirable artifacts commonly seen in the outputs of other methods. These results highlight the capability of UID-KAT to generalize effectively across diverse scenes, establishing its advantage in real-world dehazing scenarios.

\begin{figure*}
    \centering
    \includegraphics[width=1.00\textwidth]{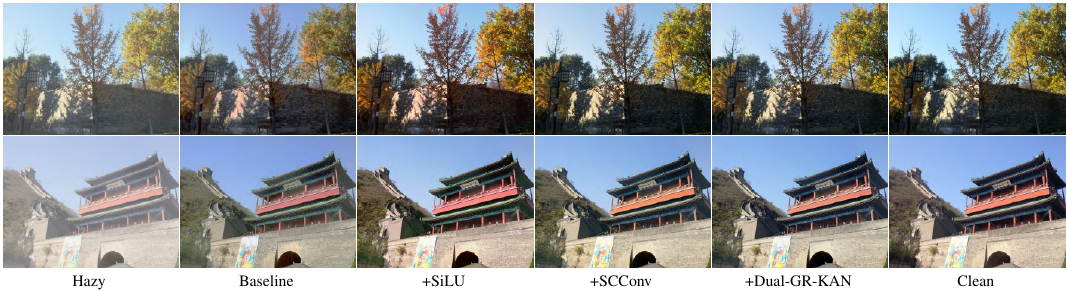}
    \caption{Effects of modifications to the baseline.}
    \label{fig:ablation_components} 
\end{figure*}

\begin{figure*}
    \centering
    \includegraphics[width=1.00\textwidth]{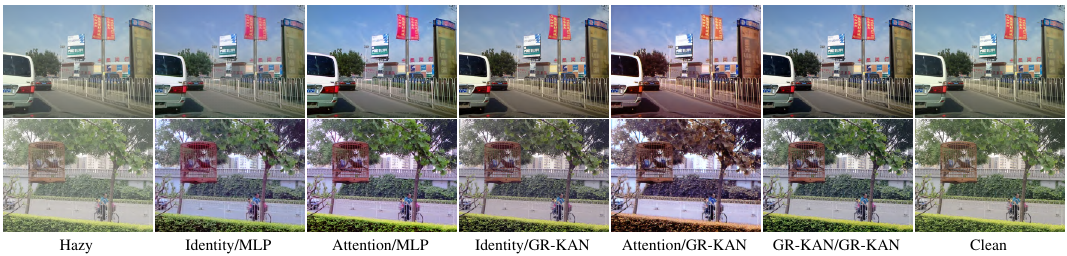}
    \caption{Effects of different mixer configurations (token mixer/channel mixer).}
    \label{fig:ablation_mixer} 
\end{figure*}

\subsection{Ablation Study}
\label{subsec:ablationstudy}

Ablation studies were carried out to determine the contribution of each component to the framework's overall performance. A standard ResNet generator was employed as the baseline, and performance on the SOTS-Outdoor and HSTS datasets was assessed by incrementally incorporating each component into the baseline model.

\begin{table}[t]
\centering
\caption{Effects of modifications to the baseline.}
\resizebox{0.47\textwidth}{!}{
\begin{tabular}{lccccrr}
\toprule
\multirow{2}{*}{\text{Backbone}} & \multicolumn{2}{c}{\text{SOTS-Outdoor}} & \multicolumn{2}{c}{\text{HSTS}} & \multirow{2}{*}{\text{\#Param}} & \multirow{2}{*}{\text{MACs}} \\
\cmidrule(r){2-3} \cmidrule(r){4-5}
& \text{PSNR} & \text{SSIM} & \text{PSNR} & \text{SSIM} & \text{(M)} & \text{(G)} \\
\midrule
Baseline & 23.01 & 0.894 & 23.71 & 0.890 & 4.59 & 17.19 \\
+SiLU & 23.14 & 0.899 & 24.07 & 0.901 & 4.59 & 17.19 \\
+SCConv & 23.22 & 0.901 & 24.56 & 0.903 & 4.88 & 20.20 \\
+Dual-GR-KAN & 24.31 & \textbf{0.917} & \textbf{26.06} & \textbf{0.922} & 7.70 & 10.59 \\
\bottomrule
\end{tabular}}
\label{table:ablation_components}
\end{table}

\begin{table}[t]
\centering
\caption{Effects of different mixer configurations (T-Mix: token mixer, C-Mix: channel mixer).}
\resizebox{0.485\textwidth}{!}{
\begin{tabular}{lccccrr}
\toprule
\multirow{2}{*}{\text{T-Mix/C-Mix}} & \multicolumn{2}{c}{\text{SOTS-Outdoor}} & \multicolumn{2}{c}{\text{HSTS}} & \multirow{2}{*}{\text{\#Param}} & \multirow{2}{*}{\text{MACs}} \\
\cmidrule(r){2-3} \cmidrule(r){4-5}
& \text{PSNR} & \text{SSIM} & \text{PSNR} & \text{SSIM} & \text{(M)} & \text{(G)} \\
\midrule
Identity/MLP & 20.86 & 0.828 & 21.35 & 0.823 & 6.51 & 10.58 \\
Attention/MLP & 22.58 & 0.884 & 23.41 & 0.883 & 7.11 & 10.59 \\
Identity/GR-KAN & 23.77 & 0.903 & 23.67 & 0.895 & 6.51 & 10.58 \\
Attention/GR-KAN & 20.20 & 0.818 & 20.51 & 0.815 & 7.11 & 10.59 \\
Dual-GR-KAN & \textbf{24.31} & \textbf{0.917} & \textbf{26.06} & \textbf{0.922} & 7.70 & 10.59 \\
\bottomrule
\end{tabular}}
\label{table:ablation_mixers}
\end{table}

\begin{table}[t]
\label{table:ablation_num_grkan}
\centering
\caption{Effects of different numbers of GR-KAN blocks.}
\resizebox{0.44\textwidth}{!}{
\begin{tabular}{lccccrr}
\toprule
\multirow{2}{*}{\text{Type}} & \multicolumn{2}{c}{\text{SOTS-Outdoor}} & \multicolumn{2}{c}{\text{HSTS}} & \multirow{2}{*}{\text{\#Param}} & \multirow{2}{*}{\text{MACs}} \\
\cmidrule(r){2-3} \cmidrule(r){4-5}
& \text{PSNR} & \text{SSIM} & \text{PSNR} & \text{SSIM} & \text{(M)} & \text{(G)} \\
\midrule
$1\times$GR-KAN & 20.71 & 0.872 & 23.56 & 0.834 & 6.51 & 10.58 \\
$2\times$GR-KAN & \textbf{24.31} & \textbf{0.917} & \textbf{26.06} & \textbf{0.922} & 7.70 & 10.59 \\
$3\times$GR-KAN & 23.57 & 0.890 & 24.73 & 0.892 & 8.89 & 10.60 \\
$4\times$GR-KAN & 23.35 & 0.892 & 24.55 & 0.892 & 10.08 & 10.60 \\
\bottomrule
\end{tabular}}
\end{table}

\textbf{Main modifications to the baseline.} This experiment verifies the impacts of key components, including SiLU, SCConv, and Dual-GR-KAN Transformer on the baseline, with the quantitative and qualitative results presented in Table \ref{table:ablation_components} and Fig. \ref{fig:ablation_components}, respectively. Initially, the default ReLU activation in the encoder and decoder is replaced with SiLU, resulting in a slightly improved performance. Next, SCConv is added to each convolution stage in both the encoder and decoder, leading to a performance gain with only a marginal increase in computational complexity. A notable improvement is observed when the proposed Dual-GR-KAN Transformer is incorporated into the feature transformation stage of the generator, validating the effectiveness of the proposed approach. These results demonstrate that the proposed modifications yield substantial improvements in both pixel-level and structural recovery, resulting in high-quality dehazing results.

\begin{table}[t]
\centering
\caption{Effects of different activation functions.}
\resizebox{0.315\textwidth}{!}{
\begin{tabular}{lcccc}
\toprule
\multirow{2}{*}{\text{Activations}} & \multicolumn{2}{c}{\text{SOTS-Outdoor}} & \multicolumn{2}{c}{\text{HSTS}} \\
\cmidrule(r){2-3} \cmidrule(r){4-5}
& \text{PSNR} & \text{SSIM} & \text{PSNR} & \text{SSIM}  \\
\midrule
ReLU & 24.13 & 0.905 & 25.52 & 0.906 \\
LReLU & 24.21 & 0.910 & 25.66 & 0.918 \\
GeLU & 23.94 & 0.871 & 24.04 & 0.893 \\
SiLU & \textbf{24.31} & \textbf{0.917} & \textbf{26.06} & \textbf{0.922} \\
\bottomrule
\end{tabular}}
\label{table:ablation_activations}
\end{table}

\begin{table}[t]
\centering
\caption{Effects of different patch sizes.}
\resizebox{0.42\textwidth}{!}{
\begin{tabular}{cccccrr}
\toprule
\multirow{2}{*}{\text{Patch size}} & \multicolumn{2}{c}{\text{SOTS-Outdoor}} & \multicolumn{2}{c}{\text{HSTS}} & \multirow{2}{*}{\text{\#Param}} & \multirow{2}{*}{\text{MACs}} \\
\cmidrule(r){2-3} \cmidrule(r){4-5}
& \text{PSNR} & \text{SSIM} & \text{PSNR} & \text{SSIM} & \text{(M)} & \text{(G)} \\
\midrule
$4\times4$ & \textbf{24.31} & \textbf{0.917} & \textbf{26.06} & \textbf{0.922} & 7.70 & 10.59 \\
$8\times8$ & 24.13 & 0.912 & 25.98 & 0.921 & 21.91 & 10.59 \\
$16\times16$ & 22.63 & 0.889 & 22.32 & 0.876 & 43.35 & 10.59 \\
\bottomrule
\end{tabular}}
\label{table:ablation_patchsizes}
\end{table}

\textbf{Mixer configurations.} The initial attempt was to apply attention for token mixing and GR-KAN for channel mixing, as proposed in \cite{yang2024kolmogorov}. However, it has been found that integrating attention into GANs for an unpaired and ill-posed task like dehazing often leads to unstable training and reduced generalizability. To investigate this further, experiments were conducted with different configurations of token and channel mixers, individually incorporating Identity, Attention, MLP, and GR-KAN and evaluating the outcomes. As shown in Table \ref{table:ablation_mixers}, the Dual-GR-KAN Transformer, which employs GR-KAN for both token and channel mixing, demonstrates a more stable training and improves the model's overall performance.

\begin{figure}[t]
  \centering
  \includegraphics[width=0.60\linewidth]{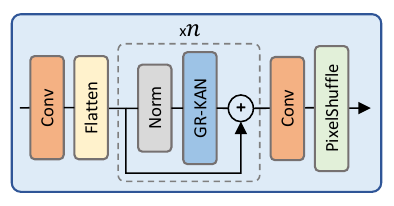}
  \caption{Abstracted architecture of GR-KAN transformer.}
  \label{fig:nxgrkan}
\end{figure}

\textbf{Number of GR-KAN blocks.} The dual-layer GR-KAN configuration has demonstrated considerable promise, prompting a natural question: What if this approach is extended to an $n \times$GR-KAN configuration? To explore this, experiments were conducted to evaluate the impacts of varying values of $n$, where $n = \{1,2,3,4\}$ represents the number of GR-KAN blocks in each level of the transformation stage, as depicted in Fig. \ref{fig:nxgrkan}. Surprisingly, these configurations are less effective as compared to the Dual-GR-KAN setting ($n=2$). This may stem from two potential reasons. First, altering the number of mixer blocks could disrupt the foundational architecture of ViTs, whose design is a well-established standard known for its effectiveness \cite{yu2022metaformer}. Second, while using only a single GR-KAN block seems insufficient, increasing the number of GR-KAN blocks appears to add redundancy, failing to enhance performance. Despite that, there may be scenarios where stacking multiple GR-KAN layers proves effective, making this a promising research direction.

\textbf{Activations.} Since the latent feature transformation stage is implemented through a series of Dual-GR-KAN Transformers, this evaluation focuses on the effect of different activation functions within the main blocks of the encoder and decoder. As summarized in Table \ref{table:ablation_activations}, LeakyReLU (0.1), referred to as LReLU \cite{maas2013rectifier}, slightly outperforms standard ReLU, resulting in improvements in PSNR and SSIM across both datasets. This can be attributed to LReLU's ability to mitigate neuron deactivation by permitting small negative slopes, which enhances feature extraction, particularly in darker regions. In contrast, GeLU \cite{hendrycks2016gaussian} shows a performance drop compared to both ReLU and LReLU. This phenomenon suggests that GeLU is less effective in low-level feature extraction, especially in image dehazing tasks, due to its non-monotonicity, which can lead to gradient reversal issues, also observed in \cite{song2023vision}. SiLU, on the other hand, achieves the best results across both datasets, benefiting from its smooth non-linearity. This helps the network model complex relationships more effectively. These outcomes highlight the superior performance of SiLU in capturing intricate patterns and enhancing feature propagation, thanks to its smooth gradient and self-gating properties.

\textbf{Patch sizes.} The influence of different patch sizes, \{$4 \times 4$, $8 \times 8$, and $16 \times 16$\}, on the framework's performance has also been analyzed. As shown in Table \ref{table:ablation_patchsizes}, the $4 \times 4$ patch size delivers the best performance across both datasets by capturing finer details, whereas increasing the patch size to $8 \times 8$ and $16 \times 16$ results in a slight and a significant drop of performance, respectively. This degradation can be attributed to the loss of critical local features, which are essential for accurately reconstructing hazy images. In terms of computational efficiency, the $4 \times 4$ patch size requires the fewest parameters (7.70M) while maintaining the same MACs (10.59G) as the larger patch sizes, making it the most efficient choice without compromising dehazing quality. Based on these analyses, the $4 \times 4$ patch size is selected as the optimal choice for the proposed framework.

\begin{table}
  \caption{Average runtime of various methods tested on CPU (Intel(R) Xeon(R) Gold 6134 @ 3.20GHz) and GPU (GeForce GTX TITAN X).}
  \centering
  \resizebox{0.36\textwidth}{!}{
  \begin{tabular}{lcc}
    \toprule
    Method & Framework & Runtime (sec.) \\
    \midrule
    DCP \cite{he2010single} & Python (CPU) & 0.114 \\
    BCCR \cite{meng2013efficient} & Python (CPU) & 0.436 \\
    RSVT \cite{tran2024single} & Python (CPU) & 0.230 \\
    YOLY \cite{li2021you} & PyTorch (GPU) & 9.324 \\
    USID-Net \cite{li2022usid} & PyTorch (GPU) & 0.039 \\
    SGDRL \cite{jia2024self} & PyTorch (GPU) & 0.032 \\
    UCL-Dehaze \cite{wang2024ucl} & PyTorch (GPU) & 0.095 \\
    \midrule
    UID-KAT-T & PyTorch (GPU) & \textbf{0.017} \\
    UID-KAT-S & PyTorch (GPU) & \textbf{0.020} \\
    UID-KAT-B & PyTorch (GPU) & \textbf{0.024} \\
    \bottomrule
  \end{tabular}}
  \label{table:runtime}
\end{table}

\subsection{Runtime}
\label{subsec:runtime} 

The efficiency of the comparison methods has been evaluated by calculating the average running time of each method on the HSTS dataset. As summarized in Table \ref{table:runtime}, traditional methods such as DCP, BCCR, and RSVT, implemented on CPU, show relatively short processing time, with DCP being the fastest at 0.114 seconds. Deep learning-based methods executed on GPU, such as YOLY, USID-Net, SGDRL, and UCL-Dehaze, exhibit significant variability in runtime. YOLY has the longest runtime of 9.324 seconds, reflecting its heavy computational demands, as this method iteratively processes an image; while other methods like USID-Net and SGDRL achieve faster execution time, with 0.039 and 0.032 seconds, respectively. On the other hand, all three variants of UID-KAT demonstrate superior efficiency, with 0.017, 0.020, and 0.024 seconds for the tiny, small, and base versions, respectively. These results indicate that UID-KAT offers an excellent trade-off between runtime and performance.

\section{Conclusions}
\label{sec:conclusions}

This paper proposes UID-KAT, a framework which leverages the capabilities of the Kolmogorov-Arnold Network (KAN) for unpaired image dehazing tasks. UID-KAT employs a newly introduced Dual-GR-KAN Transformer module, which is designed to fully exploit KAN's capabilities in transforming latent features. This innovative design enhances feature representation and effectively addresses the challenges of unpaired and ill-posed tasks like image dehazing. Furthermore, UID-KAT incorporates contrastive learning to guide the model in learning robust representations of both hazy and clean images, encouraging the model to focus on salient features and suppress irrelevant information. Experimental results show that UID-KAT achieves state-of-the-art performance across multiple benchmarks and real-world captured data, surpassing existing methods in both effectiveness and efficiency. This research highlights the promise of KANs and paves the way for future research, such as extending the framework to other ill-posed image translation tasks and further exploring the potential of KAN with advanced architectures.

\ifCLASSOPTIONcaptionsoff
  \newpage
\fi

\bibliographystyle{IEEEtran}
\footnotesize
\bibliography{bibtex/bib/ref.bib}

\end{document}